# Evaluating the Vulnerabilities in ML systems in terms of adversarial attacks

John Harshith[1**]
johnharshith@icloud.com
https://orcid.org/0000-0003-2448-6386
Mantej Singh Gill[2]
mantej.gill@gmail.com
https://orcid.org/0000-0002-2823-2415
Madhan Jothimani[3]
madhanj1999@gmail.com
https://orcid.org/0000-0002-5539-8349

1 VIT University, Vellore Campus, Vellore -
632 014, Tamilnadu, India
2 Hewlett Packard Enterprise
3 Anna University, Chennai - 600 025,Tamilnadu

**Abstract.** There have been recent adversarial attacks that are difficult to find. These new adversarial attacks methods may pose challenges to current deep learning cyber defense systems and could influence the future defense of cyberattacks. The authors focus on this domain in this research paper. They explore the consequences of vulnerabilities in AI systems. This includes discussing how they might arise, differences between randomized and adversarial examples and also potential ethical implications of vulnerabilities. Moreover, it is important to train the AI systems appropriately when they are in testing phase and getting them ready for broader use.

## Introduction

Recent progress in ML and deep learning has led to the development of highly effective models used in image classification, machine translation, game playing, and many other practical problem domains. [11, 2, 16]. Though these models demonstrate considerable performance in classification tasks, they are susceptible to adversarial inputs which confound models and lead to inaccurate predictions. Neural networks and similar classes of ML models seem to exhibit particular vulnerability to these adversarial examples, which, concerningly, can be constructed with perturbations subtle enough that they may be completely imperceptible to humans.

Though adversarial examples can take on many forms depending on the classification system, for the purposes of this paper we focus on image classification systems and the generation of adversarial images (think optical illusions for computer vision models). To formalize the problem of adversarial examples, we follow the lead of [12]. Consider an ML model $M$ that takes an input $X$ and generates a correct class prediction $y_{true}$: $M(X) = y_{true}$. It is possible to generate an adversarial input $A$ that is nearly indistinguishable from $X$, but yields an incorrect class prediction: $M(A) \neq y_{true}$. Though $A$ may be generated with the addition of only a small amount of noise to *X*, the model may be highly confident in its incorrect class prediction.

As ML models have become more ubiquitous in their application, understanding their vulnerabilities and seeking to mitigate them is increasingly necessitated. In certain critical problem domains such as health care or autonomous vehicle navigation, it is completely conceivable that such models might make ethically challenging decisions that directly affect human lives. These domains highlight the importance of developing a deeper understanding of how models generate decisions and what their shortcomings may be. With this in mind, exploration of both adversarial attacks and proposed defenses is unavoidable if we are to employ ML models confidently in ethically fraught domains.

To empirically and qualitatively investigate the properties of adversarial examples, we generate adversarial images against the pre-trained ImageNet Inception v3 system (see **Figure 1**), a state-of-the-art convolutional neural network model from Google [21]. Using a pre-trained model facilitates faster experimentation and provides us with a classification system that outperforms any model we could train ourselves on a reasonable timeline. We experiment with limiting perturbation magnitude and learning rate in multiple kinds of attacks against Inception v3. Though our experiments focus on a setting in which the attacker has full access to a model's parameters when generating adversarial examples, other work has demonstrated both the transferability of adversarial examples between models as well as techniques that can be executed against black box systems [17, 18, 7]. The property of transferability renders the conclusions of our specific research setting applicable to other black box models. After discussing the nature of these adversarial examples, we delve into possible defenses and ethical concerns surrounding adversarial examples in the context of AI safety.

The paper proceeds as follows: Section one describes related work and the current research environment. Section two formalizes the adversarial attack problem, classifies domains in which attacks may occur, and outlines methods used for generating adversarial examples. Section three presents experimental results surrounding generated examples. Section four introduces defense mechanisms and describes the implications of these adversarial examples as they relate to AI safety and security. Section five concludes the study and presents various directions for future work.

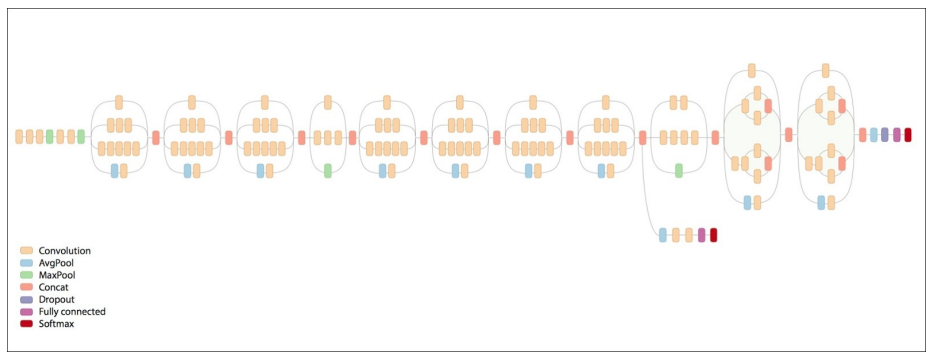

**Fig. 1.** Inception v3 model architecture from Google.

# 1 Related Work

The examination of adversarial attacks against AI classifiers began over a decade ago with naive Bayes models [6]. It wasn't until the proliferation of neural networks and deep learning, however, that adversarial examples caught the attention of the technical community. As these deeper, more complex models began to enter the mainstream, the study of their vulnerabilities accelerated. [22] was the first work to identify the seemingly widespread susceptibility of several state-ofthe-art models to adversarial examples, and began a discussion into the fragile nature of the decision boundaries exhibited by these supposedly high-performing systems. [8] expanded on this research and developed the "fast gradient sign method" for quickly generating adversarial examples given full knowledge of a model's implementation and parameters.

The property of transferability between models was then examined by [17] which opened the door to the study of adversarial attacks against black box systems in which only outputs, not parameters, were available. [12] illustrated attacks in fully black box settings where models were hosted by third parties and began the transition of research into more real world domains. [18] demonstrated the robustness of adversarial examples in real world applications by feeding physical examples through a cell phone camera before classification. This work also introduced an effective iterative approach to generating both targeted and non-targeted adversarial examples. [7] established a new general attack algorithm called "Robust Physical Perturbations" that they used to modify street signs which fooled a classifier from multiple angles and distances. This represents the current state-of-the-art in adversarial attack vectors, and its robustness certainly raises concerns about models employed in the field today and going forward.

On the defensive side, aforementioned work from [18] explored the idea of using adversarial examples during training to improve resilience against these types of attacks. They also demonstrated the added effects of adversarial training in network regularization. Gradient masking techniques and defensive distillation, which focus on hiding information from attackers through deployment considerations, were formalized by [20]. These techniques make attacks more difficult, but models will continue to be vulnerable against the same types of attacks when executed with greater computing power. Finally, [19] established an overview of the defensive landscape, discussed the trade-offs between model accuracy and resilience, and began to situate these types of attacks within the AI safety discourse. That said, there is still a dearth of research in the spheres of defense and ethics in the adversarial attack space.

# 2 Adversarial Attacks

When reviewing adversarial attacks in ML, there are multiple attack environments and vectors that must be considered. As stated, discussion in this paper will focus on attacks in the computer vision domain. In this section, we provide an overview of adversarial attack environments and vectors, then we introduce mathematical formalizations for different adversarial methods, and finally we evaluate each adversarial method and examples of their generated attacks.

## 2.1 Attack Environments

There are two primary situations in regard to model information availability: "full knowledge," where a system's architecture and parameters are accessible, and "black box," where only network outputs are accessible. The full knowledge environment is by far the most dangerous, but it is also the least likely attack vector in deployment situations with effective security practices. In this setting, an attacker knows the architecture of the underlying system and has access to the model's parameters. Parameter access allows the construction of the error gradient which can be used to directly generate adversarial examples. These examples prey upon the weakest sections of the decision manifolds and can be constructed with the least amount of noise and visual perturbation. These methods are explored in the following subsection.

The more likely environment in which attacks are to occur is the black box setting. In this situation, the model and its parameters are hidden, but its outputs are available. For example, an attacker might be able to upload an image to a computer vision system that returns information about the model's class predictions and corresponding likelihoods. Though no error gradient is available, iterative probing of the network using adversarial examples can lend directional insight into the underlying decision boundaries. As demonstrated by [17], the property of transferability also allows for the training of a similar "surrogate" model to the target model which can be used to estimate the target gradient and speed up adversarial attack generation.

Aside from considerations regarding information availability, there are also two primary types of input environments: software, where information is passed directly to a system (e.g. a picture is uploaded to a publicly available API), and physical, where a system processes information from the real world (e.g. a stop sign has been modified with adversarial stickers to mislead an autonomous vehicle). The primary distinction between these cases is environmental sterility. In the real world scenario, the attack must be physically manufactured and placed in the vicinity of the system. In the context of computer vision, such an attack must be robust across multiple viewing angles and distances. No such difficulties apply to the software scenario. That said, real world attacks are certainly feasible and improving in effectiveness [18, 7].

Finally, there are two principal attack types that can be executed against multinomial classifiers: non-targeted and targeted. In non-targeted attacks, the attacker seeks to non-directionally reduce the probability of a model producing the correct class output. An example of such an attack can be seen in **Figure 4**. The only objective of this type of attack is to reduce the probability of the current class, which in turn increases the probability of alternative classes randomly. In models where there are many potential output classes, such as in Inception v3 which has 1000 possible output predictions, non-targeted attacks tend to be less interesting [21]. Due to the similar nature of certain classes, a non-targeted attack may lead to an image of a dog being misclassified as another similar breed of dog [12]. This phenomenon led to the development of the more focused, and perhaps more nefarious, targeted attack. In this setting, a specific alternate class is selected for which to optimize prediction probability. An example of this attack carried out can be seen in **Figures 5**, **6**, and **7**.

## 2.2 Generating Adversarial Examples

This section will provide a technical overview of how adversarial examples are generated in the full knowledge environment. One should note that the techniques detailed herein provide no guarantees over whether a generated image will be misclassified by a targeted ML system. Nonetheless, these images are denoted "adversarial." This paper employs the following notation, largely informed by the approach of [12]:

- $X$: an input image, represented as a tensor along the dimensions of width, height, and depth.
- $y_{true}$: the correct class label.
- $C(X,y)$: the neural network's cost function for an input $X$ and class prediction $y$. If a network outputs a softmax distribution across classes and uses a crossentropy cost function, the cost will be equal to the negative log-likelihood of the correct class: $C(X,y) = -\log p(y|X)$.
- $Clamp_{X,\epsilon}\{X\}$: a function that clamps the pixel values of $X$, ensuring that adversarial example $X$ pixel values are within $\epsilon$ of the original image $X$, where $\epsilon$ is a modifiable hyperparameter. This limits the added noise and ensures the adversarial example is nearly visually identical to the input. The function is defined as follows:

$$Clamp_{X,\epsilon}\{X\}(x,y,z) = min(X(x,y,z)+\epsilon,max(0,X(x,y,z)-\epsilon,X(x,y,z))) \text{ where } X(x,y,z)$$

refers to the $z$ channel's value at pixel location $(x,y)$.

*Fast Gradient Sign Method* Originally introduced by [8], this method requires only a single back propagation call to retrieve an error signal and assumes a relatively linear cost function. It is less precise and generates successful adversarial examples with less subtlety than following methods but can be calculated rapidly.

$$X^{adv} = X + \epsilon \text{sign}(\nabla_X C(X,y_{true}))$$

*Iterative Non-targeted Method* This is an iterated extension of the fast method where an adversarial example is repeatedly generated and clamped at each step. The gradient error applied at each step is modulated by a learning rate $\alpha$:

$$X_0^{adv} = X, \qquad X_{N+1}^{adv} = Clamp_{X,\epsilon}(X_N^{adv} + \alpha \text{sign}(\nabla_X C(X_N^{adv}, y_{true}))$$

The number of iterations run during experimentation was balanced to allow for fast generation times as well as interesting results.

*Iterative Targeted Method* This is a modified version of the iterated method and a slight modification of the iterative least-likely class method devised by [12]. Instead of just increasing the error of the originally predicted class, this method seeks to decrease the error of a specific selected class label $y_{target}$ [10]. To generate an adversarial example, method maximizes $\log(p(y_{target}|X))$[24] by transforming the image in the direction of $\text{sign}\nabla_X \log(p(y_{target}|X))$.

$$X_0^{adv} = X, \qquad X_{N+1}^{adv} = Clamp_{X,\epsilon}(X_N^{adv} - \alpha \text{sign}(\nabla_X C(X_N^{adv}, y_{target}))$$

The same hyperparameters and number of iterations can be employed as in the non-targeted approach.

## 3 Experimental Results

### 3.1 Experimental Design

To compare the capabilities of each adversarial example generation method, we examine the performance of each method on a subset of ImageNet images across a range of $\epsilon$ values. Intuitively, an adversarial image generated with $\epsilon = 0$ yields the same image as before. A higher $\epsilon$ indicates more leeway for modification of the original image. The learning rate $\alpha$ is held constant at 1 across all experiments. To examine the targeted method, we choose a random target class to optimize for. Although there is a chance that this random class will lay close to the true class, this is rather unlikely across 1000 possible outputs. Due to the computationally expensive nature of generating adversarial examples with the iterative methods, we unfortunately had to resort to a relatively small sample size at each value of $\epsilon$, quantitatively evaluating each method on a subset of only 20 random samples. We also perform a qualitative analysis of each method's performance and behavioral tendencies by examining generated adversarial images.

0.49

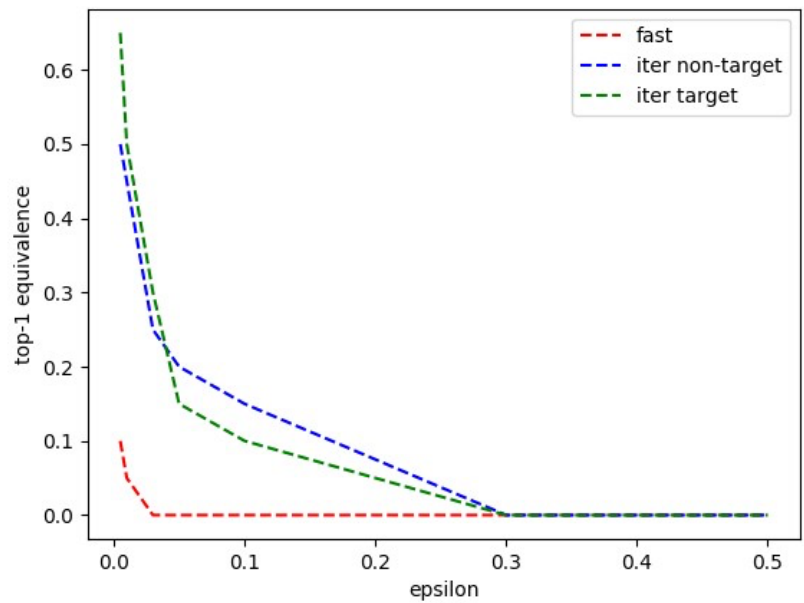

0.49

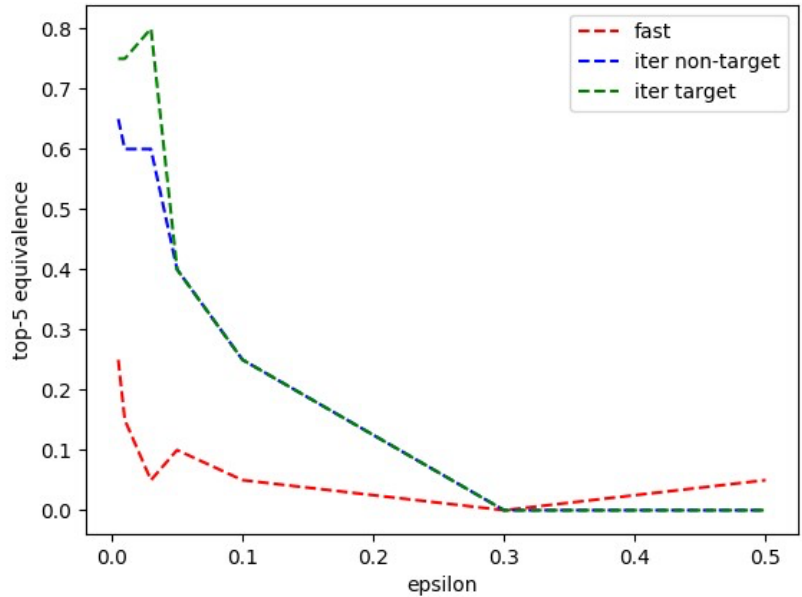

**Fig. 2.** Top-1 and top-5 model accuracy when processing adversarial examples. The fast gradient sign, iterative non-targeted, and iterative targeted methods are compared for different values of $\epsilon$.

### 3.2 Experimental Results

To evaluate performance, we compare the model's predictions for adversarial images against its predictions for unmodified images. As mentioned above, there are no guarantees on whether an adversarial image generated with these methods will successfully fool the model. First, each method was evaluated on its ability to alter the model's top-1 accuracy. We define top-1 accuracy to be the rate at which the model's top prediction for each initial image matches its top prediction for each adversarial image. Next, we evaluated each method on its ability to alter the model's top-5 accuracy. We define top-5 accuracy to be the rate at which the model's top prediction for each initial image appears within its five top predictions for each adversarial image. Results for each adversarial method's effectiveness at different values of $\epsilon$ can be seen in **Figure 2**.

The motivation for examining both top-1 and top-5 accuracy lies in the difference between the targeted and non-targeted methods. The targeted method seeks to increase the likelihood of a random of alternative class. As discussed, it is likely that this class is far away from the initially predicted class. Because of this distance, the error gradient will pull the adversarial image away from all the top classes (which should be grouped together in the class space) faster. When the *non-targeted* method is run, the likelihood of the top class is reduced, and though this is turn will reduce the likelihood of classes in its neighborhood, there is less of a pull towards a completely new area in the latent class space. In summation, we might hypothesize that the targeted method would reduce the top-5 accuracy of the model more rapidly than the non-targeted method. In our limited results, this hypothesis was not confirmed as the targeted and non-targeted attacks performed very similarly.

One interesting thing to note is the apparent effectiveness of the fast gradient sign method. This is slightly misleading, however, as this method operates in a less subtle fashion than the other two by effectively introducing $\epsilon$-scaled noise in a single shot. For a given $\epsilon$, the fast method is far more "destructive" as it modifies individual pixel data more aggressively. The iterative methods attempt to create an attack smoothed over the image space, thereby reducing the visual artifacts introduced by modification. **Figure 3** visually demonstrates the destructive nature of the fast method. The swans in the image have been visibly manipulated through the introduction of foreign colors even at a relatively small $\epsilon$. **Figure 4** illustrates an iterated attack at the same $\epsilon$ that introduces far less visible noise.

**Figures 4** and **5** showcase the non-targeted and targeted adversarial methods best. Any noise introduced is remarkably subtle, even though the model's class predictions for the attacking image are completely different. In the first image, the model changes its prediction from "convertible" to "crayfish"; in the second, from "Granny Smith" apples to "cello." A human looking closely might be able to spot defects within these images, but they would never make the same severity of classification mistakes as the model. **Figures 6** and **7** are included to demonstrate the destruction introduced by the iterative algorithms at higher values of $\epsilon$. Images generated with integer values of $\epsilon$ and higher display significant artifacts. Interestingly, these adversarial images would still likely be correctly classified by a human even though the model may produce an incorrect prediction with near 100% confidence. This last point is an important one to note regarding adversarial attacks. When the model processes these adversarial images, its misclassifications are incredibly confident. In **Figure 6**, for example, the model does not "see" strange looking apples—it very confidently "sees" a cello [23].

In summary, the fast gradient sign method is effective, but more destructive and less subtle than the iterative methods. Both iterative methods performed similarly quantitatively, though in models with a large number of output classes we expect the iterative targeted approach to be more successful in reducing accuracy, especially top-5 accuracy. While generating adversarial images with iterative methods is slower, a qualitative analysis indicates their superiority in fooling the model while still retaining visual information from the original image.

## 4 Defensive and Ethical Considerations

In this section, we will discuss current research on potential defensive mechanisms that could be employed to combat adversarial attacks, as well as the ethical considerations that must be weighed before deploying systems susceptible to such attacks.

### 4.1 Defensive Approaches

When assessing the security of an ML system deployed in the wild, there are many attack vectors to consider. An adversary may attempt to gain access to the deployment of the model itself or perhaps provide malicious inputs if a model is trained on-line. We saw this latter case play out particularly poorly for Microsoft with the launch of its chatbot Tay [13]. For the purposes of this paper, we will narrow the scope to focus only on defending against attackers employing the types of subtle adversarial techniques shown thus far.

These defenses can be distilled into two primary camps: *reactive* and *proactive* [15]. An example of a reactive defense might be the preprocessing and/or sanitization of all inputs by another model. It is feasible to train a model to recognize adversarial inputs before they ever reach the primary classifier[25, 26]. This solution is far from ideal, however, as it requires the maintenance of two heavy duty models in production instead of one. It also opens up the possibility of incorrectly flagging legal inputs, which could be more limiting to the original system than the possibility of adversarial inputs in the first place. There is a certain inelegance to this approach as well; though the problem of robust vision is far from simple, the fact that humans do not fall prey to these types of schemes is encouraging to the technical community that more complete proactive solutions exist.

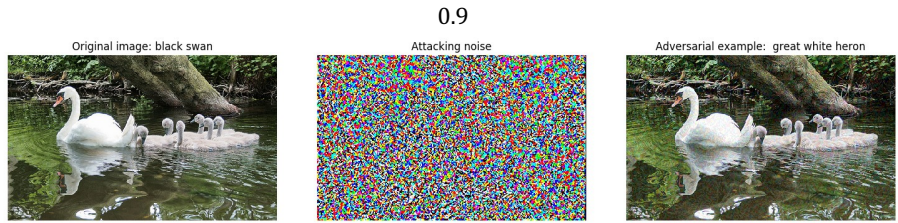

**Fig. 3.** Fast gradient sign attack with $\epsilon = 0.05$. The attack modifies the prediction, but yields a similar class.

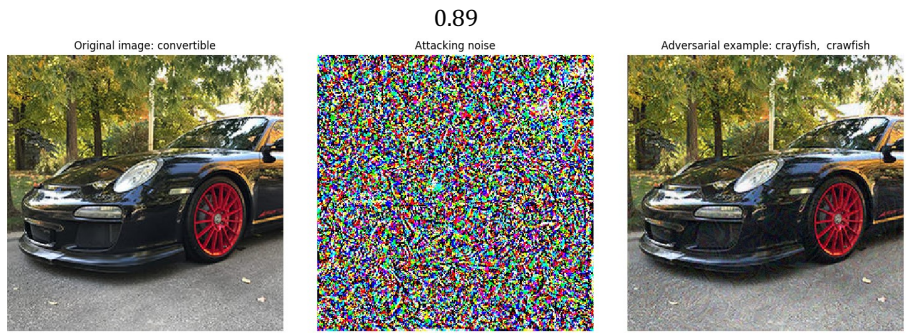

**Fig. 4.** Iterative non-targeted attack with $\epsilon = 0.05$. The attack successfully modifies the prediction.

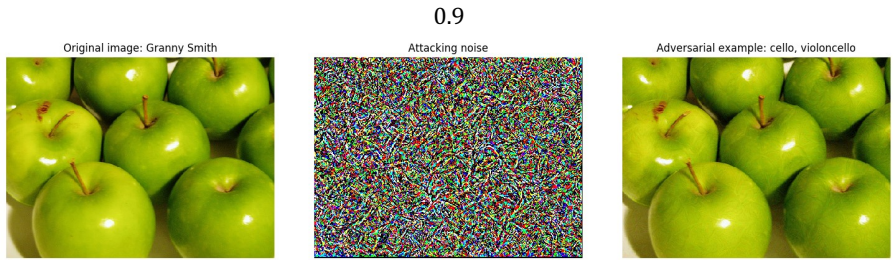

**Fig. 5.** Iterative targeted attack with $\epsilon = 0.02$. The attack was successful given the target class "cello."

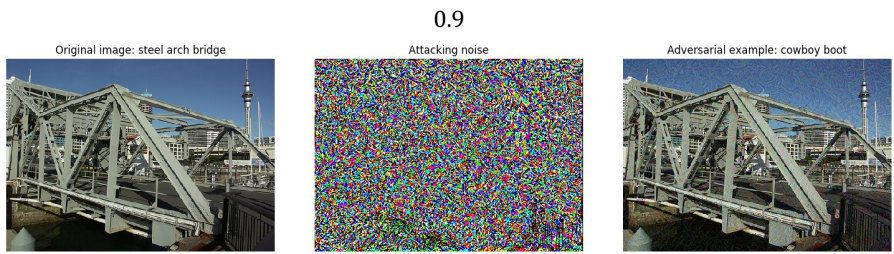

**Fig. 6.** Iterative targeted attack with $\epsilon$ = 0.08. The attack was successful given the target class "cowboy boot."

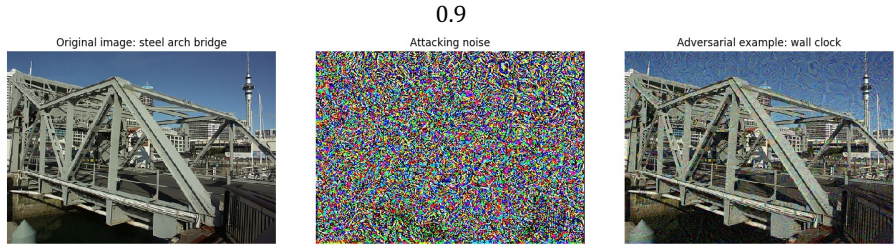

**Fig. 7.** Iterative targeted attack with $\epsilon$ = 10.0. The attack was successful given the target class "wall clock."

**Fig. 8.** Examples of adversarial images generated with different methods and $\epsilon$ values. Above each image the model's most likely class prediction is shown.

*Adversarial Training* One current leading proactive approach is that of adversarial training. During the training process, the gradients of the unfinished model are used to generate adversarial examples with the same methods described earlier. These adversarial examples are then used to augment the training dataset, thereby increasing the robustness of the model against images with these previously unnoticed perturbations [19]. Though adversarial training does improve the resilience of a model against such attacks, along with the added benefit of increased regularization, the decision boundaries remain relatively fragile. An attacker with more computing power still seems able to locate weaknesses. One drawback is the increased complexity of training; additional computing resources and time are necessary to realize this training schema. Nevertheless, the adversarial training approach is a first step in the direction of developing more durable models.

*Gradient Masking, Defensive Distillation, and Label Smoothing* Another set of approaches rely on minimizing the information available to an attacking algorithm. Though these techniques are technically different, they are similarly motivated and will thus be considered in tandem. In the process of distillation, a larger, more complicated model is compressed into a smaller form while sacrificing a very small amount predictive accuracy. The intuition behind this approach is that the smaller model will learn a "softer" probability distribution and will encode less helpful information to an attacker in its output predictions than the larger model [20]. Gradient masking similarly relies on information reduction by artificially limiting gradient norm during training, thus dramatically decreasing the signal available to an attacker at inference time. A final approach in this category is label smoothing, where the output probabilities are held closer together (i.e., there is no single class with extremely high probability); this should theoretically limit information by making it more difficult to target a specific class [19].

Though these defenses have been shown to increase the difficulty of generating adversarial examples, they are not insurmountable at this time. By the transferability property introduced by [17], an attacker can create a surrogate model given access to similar training data and basic knowledge of the target model's architecture. This surrogate model can then be used to obtain a gradient not dissimilar from that of the target, which can then in turn be used to fine tune adversarial examples [19]. This surrogate model makes the distinction between full knowledge and black box scenarios far less relevant.

## 4.2 Ethical Considerations

We have technically described the nature of adversarial attacks and the limitations of available defenses, but we have not yet made it clear why we should care about such vulnerabilities. Though ML systems have reached mass application in some domains, we are only at the beginning of the adoption curve. Many traditional software systems will be replaced by more intelligent ML algorithms, and sometimes not without cost. Despite their increased flexibility and intelligence, systems built with neural networks and similar approaches are not only susceptible to adversarial attacks, but are also increasingly opaque in their decision-making. As our society continues to place a greater quantity of increasingly complex decisions in the hands of these systems, it is concerning that our understanding of their operation seems to be decreasing.

We can conceive of highly critical applications in which adversarial examples are particularly troubling. Banks that process checks programmatically might easily be defrauded by an adversarial forgery that appears legitimate to a human. Autonomous weapon systems that rely on computer vision, as described by [1] and others, might be confused by disguised weapons or tricked into firing on innocents that are invisibly marked against their knowledge. An autonomous vehicle might be fooled into interpreting a stop sign as a 45 mph sign [7]. This last example is specifically relevant given the seeming inevitability of autonomous vehicles overtaking our streets in the coming years. Further, it is concerning that this specific attack has already been demonstrated successfully in real world settings.

Even though these models exhibit clear defects in adversarial settings, it is difficult to make a conclusive recommendation on their deployment. Improving training practices is a good place for the technical community to start, though this is easier said than done. The work of [14], which focuses on improving outcomes through the tighter integration of human oversight during training, is a positive contribution to the advancement of AI safety; unfortunately, it is not applicable when the problems encountered during training are at times more subtle than humans can perceive. Adversarial training should be encouraged in the development of these models, and it brings along the positive side effect of increased regularization. Information masking techniques, despite their provable fragility, also do increase the safety and resilience of ML systems deployed in the real world.

*Fairness* Even beyond the scope of adversarial examples, it is important to think about the fairness of employing such opaque models. In recent work evaluating the fairness of recidivism prediction systems, [4] demonstrated that bias free predictive instruments can still result in disparate impact across populations when input data is not carefully curated. Beyond augmenting ML training data with adversarial examples, it is necessary for the technical community to become more careful about dataset construction, especially in ethically complex domains like recidivism prediction. Institutions and organizations could require the use of black box auditing tools like FairML to prevent the manifestation of obvious biases in production systems.

*Accountability* As we do employ these more advanced systems, we develop an increased expectation of their accountability. This accountability is often a justification for their necessity. Despite limitations, autonomous vehicles are easy to justify when they are expected to dramatically reduce the frequency of accidents [3]. Our evaluation of the accountability of these systems, however, is framed by an assumption of generally good faith actors. The ethical calculus of autonomous vehicles and many other systems certainly changes in the face of malicious adversarial attacks. Can we permit these systems to be responsible for human lives when it is seemingly simple for a knowledgeable actor to influence them with attacks hidden in plain sight? At this current juncture given the limited scope of these systems in real world applications, there is little cause for concern. But in coming years when ML systems dictate greater portions of our lives, slight perturbations may indeed be a worthy anxiety.

*Trade-offs* The systems in development today are far from perfect, and in all likelihood they'll continue to have flaws for the remainder of their existence. It is necessary then not to write them off entirely, but to evaluate their trade-offs. One of the essential trade-offs in models vulnerable to adversarial attacks is the trade-off between representative capacity and interpretability. [9] showed that multilayer feedforward neural networks are universal approximators; essentially, a model with enough parameters can represent any function mathematically. As model size and complexity increase, however, the variety and quantity of data required to prevent overfitting increases.

At this time, our models are limited in representative capacity by available computing power and data; this opens them up to the kinds of manipulations abused by adversarial attacks. In order to combat these attacks in coming years, more complicated models will be trained on larger datasets. This practice, though, comes at the cost of interpretability. The decisions made by such models will become progressively difficult to explicate, which might prove problematic in fields like health care where the reasoning behind certain decisions should be comprehensible [5]. There is no obvious solutions to these concerns. At first, it makes sense to require that models used in such critical applications should be highly interpretable. Yet models limited by such a requirement will have lesser representative power and will in turn make less intelligent decisions. Would we be willing to sacrifice potentially worse outcomes for a better understanding of how they came about?

In total, a model should be evaluated not just on its predictive accuracy, but also along the dimensions of robustness, fairness, accountability, and interpretability. There is no free lunch when it comes to ML models, and these properties are at times orthogonal to one another. That said, the threat of adversarial attacks, especially in ethically critical domains, should be taken seriously. Defensive techniques like adversarial training should be used to improve the resilience and safety of models in the field.

## 5 Conclusion and Future Work

In this paper we have provided an overview of the current state of adversarial attack research as it pertains to machine learning models. We have reviewed several adversarial image generation methods and identified the family of iterative methods as particularly promising given their more subtle approach. Aside from a technical evaluation of attacks, we have outlined primary known defensive mechanisms alongside analysis of their shortcomings. Finally, we situate this work within the AI safety discussion and elucidate ethical concerns surrounding the deployment of models susceptible to adversarial attacks. Despite deficiencies in defensive mechanisms, we encourage the technical community to take these methods seriously to increase the safety and accountability of models employed in real world settings. Future work should focus on developing adversarial attacks and defenses together while at the same time improving model interpretability. A more nuanced understanding of how opaque ML models form decision boundaries will be vital to their successful deployment in coming years.

## 6 Author Contribution

Mantej Singh Gill devised the primary conceptualisation, research, oversight, technique, and validation and carried out the reviewing and editing part of the paper. John Harshith worked on Data curation, investigation, methodology, and visualisation and contributed to the original draft preparation. Madhan Jothimani helped in the interpretation of the results and provided critical feedback to shape the overall work. All authors discussed the results and contributed to the final manuscript.